\documentclass{article}

\usepackage{microtype}
\usepackage{graphicx}
\usepackage{subfigure}
\usepackage{booktabs} %
\usepackage{siunitx}

\usepackage{hyperref}

\usepackage[arxived]{icml2024}

\usepackage{amsmath}
\usepackage{amssymb}
\usepackage{mathtools}
\usepackage{amsthm}
\usepackage{enumitem}
\usepackage{multirow}
\usepackage{makecell}
\usepackage{graphicx}
\usepackage{subfigure}
\usepackage{amssymb}%
\usepackage{pifont}%
\newcommand{\cmark}{\textrm{\ding{51}}}%
\newcommand{\xmark}{\textrm{\ding{55}}}%
\usepackage{comment}
\usepackage[normalem]{ulem}

\usepackage[capitalize,noabbrev]{cleveref}

\theoremstyle{plain}

\theoremstyle{definition}

\theoremstyle{remark}

\usepackage[textsize=tiny]{todonotes}
\usepackage{xspace}

\newcommand{\method}{Best-fit Packing\xspace}

\begin{document}

\twocolumn[
\icmltitle{Fewer Truncations Improve Language Modeling}

\icmlsetsymbol{equal}{*}

\begin{icmlauthorlist}
\icmlauthor{Hantian Ding}{yyy}
\icmlauthor{Zijian Wang}{yyy}
\icmlauthor{Giovanni Paolini}{yyy}
\icmlauthor{Varun Kumar}{yyy}
\icmlauthor{Anoop Deoras}{yyy}
\icmlauthor{Dan Roth}{yyy}
\icmlauthor{Stefano Soatto}{yyy}
\end{icmlauthorlist}

\icmlaffiliation{yyy}{AWS AI Labs}

\icmlcorrespondingauthor{Hantian Ding}{dhantian@amazon.com}
\icmlcorrespondingauthor{Zijian Wang}{zijwan@amazon.com}

\icmlkeywords{Machine Learning, ICML}

\vskip 0.25in
]

\printAffiliationsAndNotice{}  %

\newcommand{\xxcomment}[4]{\textcolor{#1}{[$^{\textsc{#2}}_{\textsc{#3}}$ #4]}}

\newcommand{\zijian}[1]{
 \xxcomment{blue}{Z}{W}{#1}
}
\newcommand{\dhantian}[1]{
 \xxcomment{red}{H}{D}{#1}
}

\newcommand{\nsig}[1]{\textbf{#1}\textsuperscript{\textbf{n}}}
\newcommand{\sig}[1]{\textbf{#1}\textsuperscript{\textbf{s}}}

\newcommand{\nsigs}[1]{\textbf{#1}\textsuperscript{\textbf{ n}}}
\newcommand{\sigs}[1]{\textbf{#1}\textsuperscript{\textbf{ s}}}

\newcommand{\rcn}[1]{\hspace{5.5pt}\nsig{#1}}
\newcommand{\rcs}[1]{\hspace{5.5pt}\sig{#1}}

\newcommand{\nln}[1]{\hspace{4pt}\nsig{#1}}
\newcommand{\nls}[1]{\hspace{4pt}\sig{#1}}

\newcommand{\csn}[1]{\hspace{4pt}\nsig{#1}}
\newcommand{\css}[1]{\hspace{4pt}\sig{#1}}

\newcommand{\smn}[1]{\hspace{3.5pt}\nsig{#1}}
\newcommand{\sms}[1]{\hspace{3.5pt}\sig{#1}}

\newcommand{\pln}[1]{\hspace{4pt}\nsig{#1}}
\newcommand{\pls}[1]{\hspace{4pt}\sig{#1}}

\newcommand{\rclow}[1]{\hspace{-0pt}#1}

\begin{abstract}
In large language model training, input documents are typically concatenated together and then split into sequences of equal length to avoid padding tokens. Despite its efficiency, the concatenation approach compromises data integrity---it inevitably breaks many documents into incomplete pieces, leading to excessive truncations that hinder the model from learning to compose logically coherent and factually consistent content that is grounded on the complete context. To address the issue, we propose \method, a scalable and efficient method that packs documents into training sequences through length-aware combinatorial optimization. Our method completely eliminates unnecessary truncations while retaining the same training efficiency as concatenation. Empirical results from both text and code pre-training show that our method achieves superior performance (e.g., relatively +4.7\% on reading comprehension; +16.8\% in context following; and +9.2\% on program synthesis), and reduces closed-domain hallucination effectively by up to 58.3\%.

\end{abstract}

\section{Introduction}
\label{sec:intro}

\begin{figure*}[ht]
    \centering
    \includegraphics[trim=0 0 0 0, clip, width=1.0\textwidth]{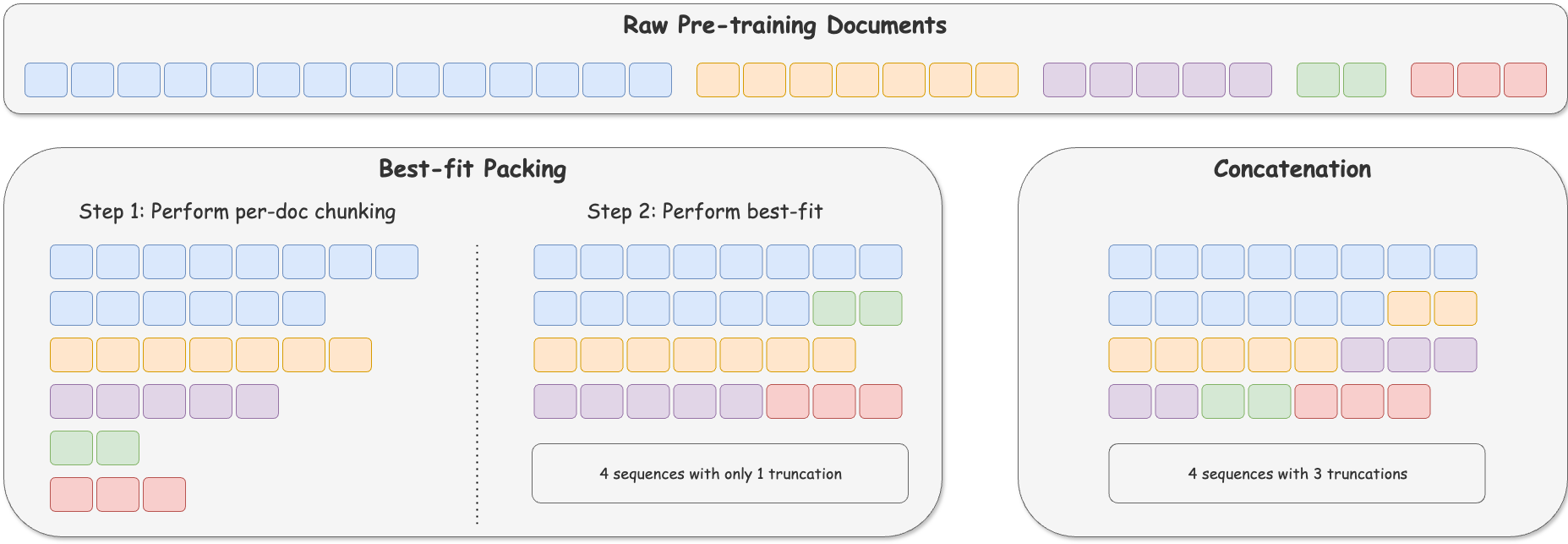} 
    \vspace{-0.18in}
    \caption{An illustration of the proposed \method compared with concatenation (baseline). We set max sequence length to 8 tokens in this example. \textbf{Top: }Original training documents. Each box stands for a token. Contiguous boxes in the same color represent a document. There are five documents of lengths 14, 7, 5, 2, 3, respectively. \textbf{Bottom-left: }\method. In step 1, we segment the long document (e.g., blue) into chunks with $\leq$ 8 tokens. In step 2, we group chunks into training sequences in a smart way that results in the smallest number of sequences. We do not break any chunk in the second step. In total, only one document was truncated and this is necessary to meet the max sequence length requirement. \textbf{Bottom-right: }The concatenation approach. 3 out of the 5 documents are truncated.}
    
    \label{fig:main}
\end{figure*}

Large language models (LLMs) have achieved unprecedented success on a number of natural language processing and coding benchmarks \cite{gpt3,codex} and in complex real-world tasks \cite{instructgpt}. This remarkable progress is driven by large-scale pre-training over a massive amount of unlabeled documents.
When formatting the training inputs, na\"{i}vely padding every document to a fixed length is inefficient as short documents lead to an excessive amount of padding. Instead, the common practice is to concatenate all documents together and then split them into sequences of exactly the model's context length. A sentinel token (e.g., \texttt{<|endoftext|>}) is often added at the end of each document to indicate document boundaries within each training sequence.
This concatenate-then-split (hereafter ``concatenation'') approach
has been widely adopted in training language models in both natural language \cite{gpt3,palm,gopher,opt,llama2,bloom} and programming language \cite{codegen}, thanks to its optimal training efficiency as no padding is needed.
However, such training efficiency comes at the expense of \textit{data integrity}---documents that could have been processed in their entirety by the model are instead fragmented into independent segments, which naturally results in \textit{loss of information}.
Further, truncation reduces the amount of context within each segment, causing next-token prediction to be potentially \textit{ungrounded} to its context, and thus making models more \textit{prone to hallucination}.

We argue that data integrity is the key towards better language modeling. To realize this, we first show that it is feasible to  group billions of documents at pre-training scale into sequences in a way that is as token-efficient as concatenation without incurring \textit{any} unnecessary truncation: only documents beyond model's context length need to be segmented. Data grouping strategies that preserve the entirety of individual samples have been widely adopted for encoder-only and encoder-decoder models \cite{roberta, t5, bert_packing}. Nonetheless, these existing strategies either exhibit limited scalability or compromise training efficiency, making them less favorable compared to the concatenation method in LLM training at scale.

In response, we propose \textit{\method} to eliminate unnecessary document truncations without sacrificing training efficiency. As illustrated in Figure \ref{fig:main}, we first segment long documents into multiple chunks by model's context length. Documents shorter than that are kept as singleton chunks. Next, we pack all the chunks into training sequences without breaking them any further. This step is essentially an instance of \textit{the bin packing problem}\footnote{Bin packing is an optimization problem in which items of different sizes must be packed into a finite number of bins or containers, each of a fixed given capacity, in a way that minimizes the number of bins used \cite{bernhard2008combinatorial}.}, which is NP-hard. We employ Best-Fit-Decreasing \cite{best-fit-earliest}, an approximation algorithm, and further optimize it to handle billions of documents efficiently. Empirical results show that the packed training sequences only contain a negligible amount of padding, which enables us to maintain the same training efficiency as the concatenation approach while preventing unnecessary document truncation.

To validate the effectiveness of truncation reduction, 
we pre-train a set of models with inputs formatted by concatenation and \method respectively, ranging from 7B to 13B in model size, 2k to 8k in sequence length, on both Natural Language (NL) and Programming Language (PL) data.
We evaluate these models on 22 tasks covering reading comprehension, natural language inference, context following, summarization, world knowledge, and program synthesis. 
Experiment results show that models trained with fewer truncations demonstrate superior performance and exhibit less hallucination.

In summary, our main contributions are the following.
\begin{itemize}%
    \item We highlight the truncation issue inherent in the widely-used concatenation method for LLM pre-training (\S\ref{sec:prelim}).
    \item We analytically show the adverse impact of truncation on learning through a simplified model (\S\ref{sec:analytical}).
    \item We propose \method, a scalable data grouping method that eliminates unnecessary document truncations at almost no cost of training efficiency  (\S\ref{sec:packing}). 
    \item We empirically quantify the benefits of truncation reduction in a variety of downstream scenarios (\S\ref{sec: results}). 
\end{itemize}

\section{The Curse of Truncation}\label{sec:prelim}

\begin{figure*}
    \centering
    \subfigure[Undefined Names\label{fig:concat_a}]{\includegraphics[width=0.3\linewidth]{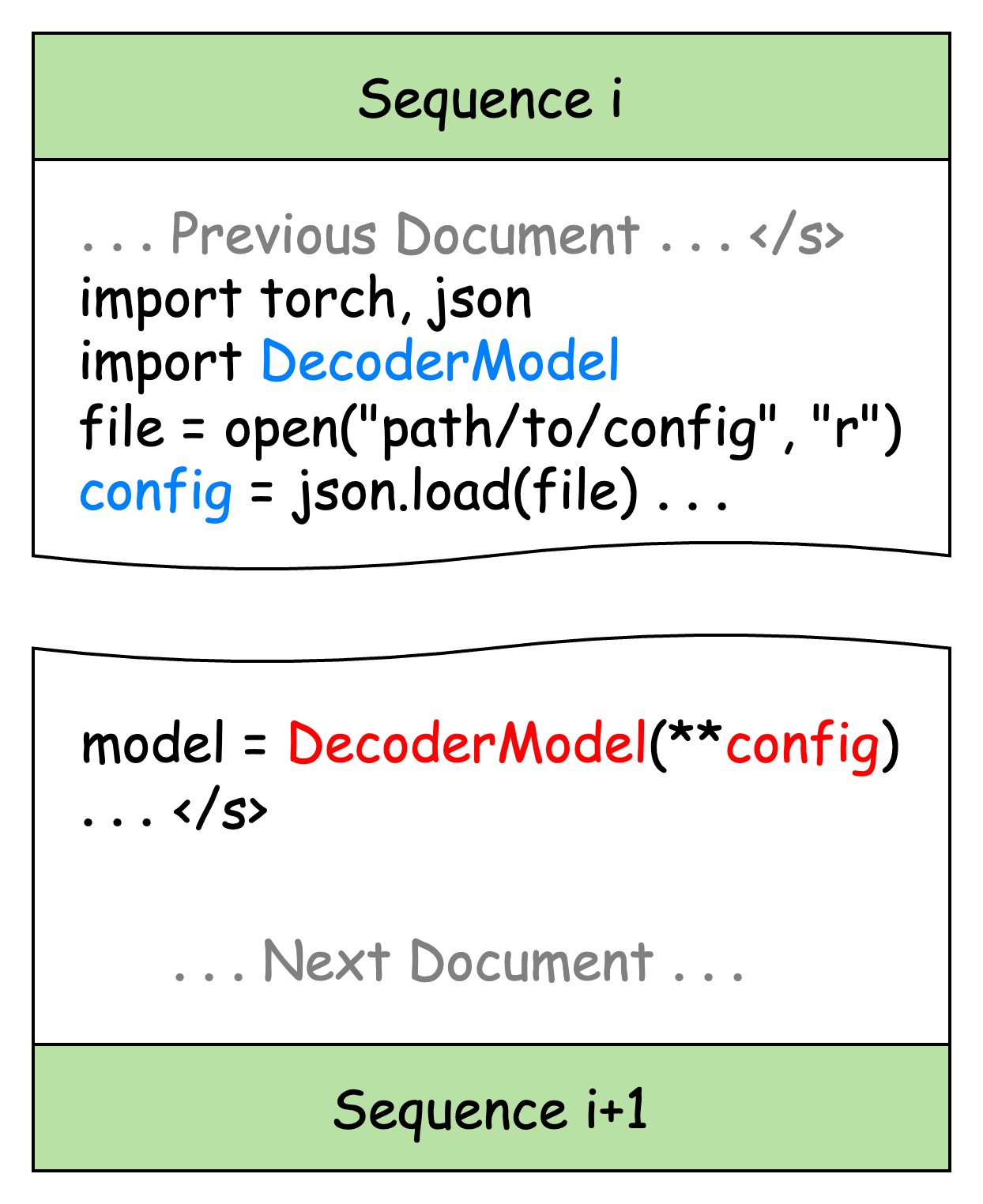}}
    \subfigure[Ungrounded content\label{fig:concat_b}]{\includegraphics[width=0.3\linewidth]{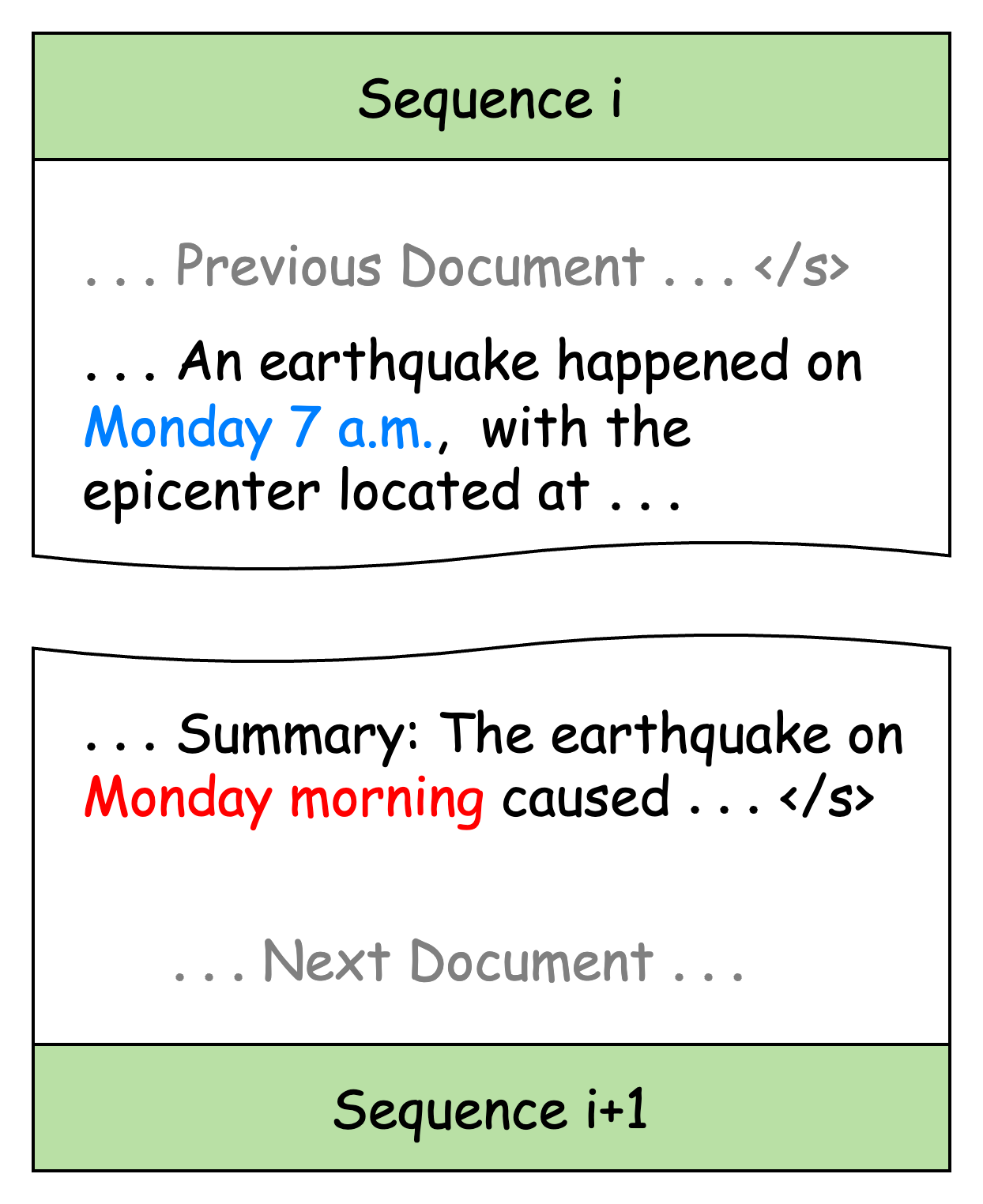}}
    \subfigure[Missing knowledge\label{fig:concat_c}] {\includegraphics[width=0.3\linewidth]{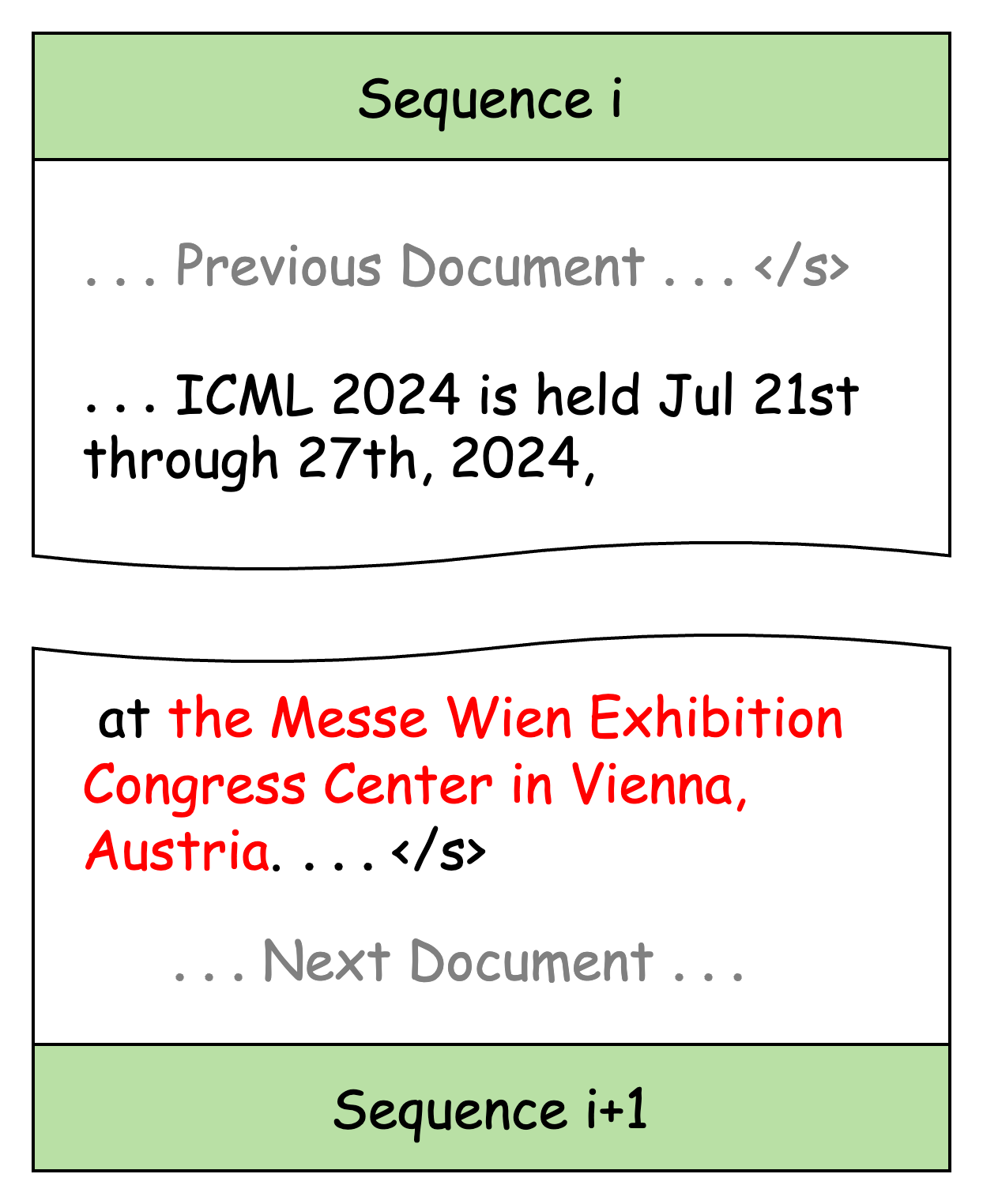}}
    \caption{Examples where document truncation leads to hallucination or loss of knowledge.
    (a) Variable definitions ({\color[HTML]{007FFF}in blue}) are truncated and subsequent usage calls result in undefined names ({\color[HTML]{FF0000}in red}).
    (b) Key context information is truncated ({\color[HTML]{007FFF}in blue}), making the summary unfaithful ({\color[HTML]{FF0000}in red}).
    (c) Where ICML 2024 is held is unknown to the model due to truncation.
    }
    \label{fig:concat_ex}
\end{figure*}

A well-written document in its entirety is naturally coherent and self-contained. In particular, factual statements in the document often logically depend on their aforementioned context through reference, entailment, or more sophisticated reasoning. We refer to the key span(s) of context that serves to establish such a dependency relation as \textit{grounding context}. When learning from next-token prediction, if the grounding context is missing, the model will be forced to spuriously predict token(s) that in fact cannot be derived from the observed partial context. Consequently, at inference time, the model has a higher chance to ignore the grounding context (even when it is provided) and generate content that either contradicts or cannot be verified from the given context, which is known as \textit{closed-domain hallucination}\footnote{\textit{Hallucination} is an overloaded term. In this work, we focus on context-based hallucination as opposed to knowledge-based.} \cite{gpt4}. We illustrate this point in Figure \ref{fig:concat_ex}.

Figure \ref{fig:concat_a} shows an example in Python. Despite the original code being correct, splitting variable definitions and corresponding usages into two distinct training sequences introduces grammatical errors. As self-attention does not cross sequence boundaries, \texttt{DecoderModel} and \texttt{config} are essentially undefined in the latter training sequence. Formatting data in such a fragmented way makes models learn pathological patterns, potentially leading to hallucination in downstream tasks. For example, in a program synthesis task, the model may directly use \texttt{config} without its definition. Even worse, the model may disregard the provided context and fabricate an irrelevant name: if we intend to instantiate an \texttt{EncoderModel}, and specify \texttt{import EncoderModel} in context, the model may still generate \texttt{model=DecoderModel(\dots)} due to the learned spurious association between \texttt{model} and \texttt{DecoderModel}.

Figure \ref{fig:concat_b} illustrates the same issue in natural language where truncation harms faithfulness. The phrase \textit{Monday morning} in the summary cannot be grounded to any part of the context in the same training sequence, and thus turns into a fabrication. Such incomplete samples can reduce models' \textit{context-awareness} (i.e., the ability to attend to context) and result in unfaithful generation or nonsensical reasoning.

Besides exacerbating hallucination, truncation can also impede knowledge acquisition during training, as textual representation of knowledge often takes the form of complete sentences or paragraphs, which is vulnerable to fragmentation. For example, in Figure \ref{fig:concat_c}, the model will not be able to learn the location of ICML because the conference name and its venue are located in different training sequences.

\subsection{Analytical Study via a Simplified Stochastic Process}
\label{sec:analytical}

As an additional source of intuition, we describe a simplified stochastic process $(X_n)_{n \in \mathbb{N}}$ for which we can analytically show that a model trained on truncated sequences achieves a \textit{strictly worse} sequence modeling accuracy than a model trained on full sequences, even if the amount of training data is infinite.
While it is difficult to rigorously establish a theory on how truncation impacts learning with transformers models, we would like to make a first attempt on the analytical exploration to better motivate our proposal.

In analogy with language modeling, we can think of the $X_n$'s as tokens in the binary vocabulary $\{0, 1\}$. Our process is defined recursively, starting from a Bernoulli variable $X_0$ which takes the value $0$ with probability $0.5$ and the value $1$ otherwise.
For $n\geq 1$, the variable $X_n$ takes the value of $X_0$ with probability $p$ and $1-X_0$ with probability $1-p$, where $p \in (0.5, 1)$ is fixed.
A graphical model associated with this process would be a tree with $X_0$ as the root and $X_1, X_2, \dots$ as the leaves.

We now compare a ``model A''  trained on sequences $X_{0:L} := (X_0, X_1, \dots, X_{L-1})$, against a ``model B'' trained on sequences $(X_0)$ and $X_{1:L} := (X_1, X_2, \dots, X_{L-1})$. Thus, training of model B is affected by truncation.
We assume that there is a sufficient amount of data for the models to perfectly fit the training sequences.

\begin{figure}[!h]
    \centering
    \includegraphics[width=0.95\columnwidth]{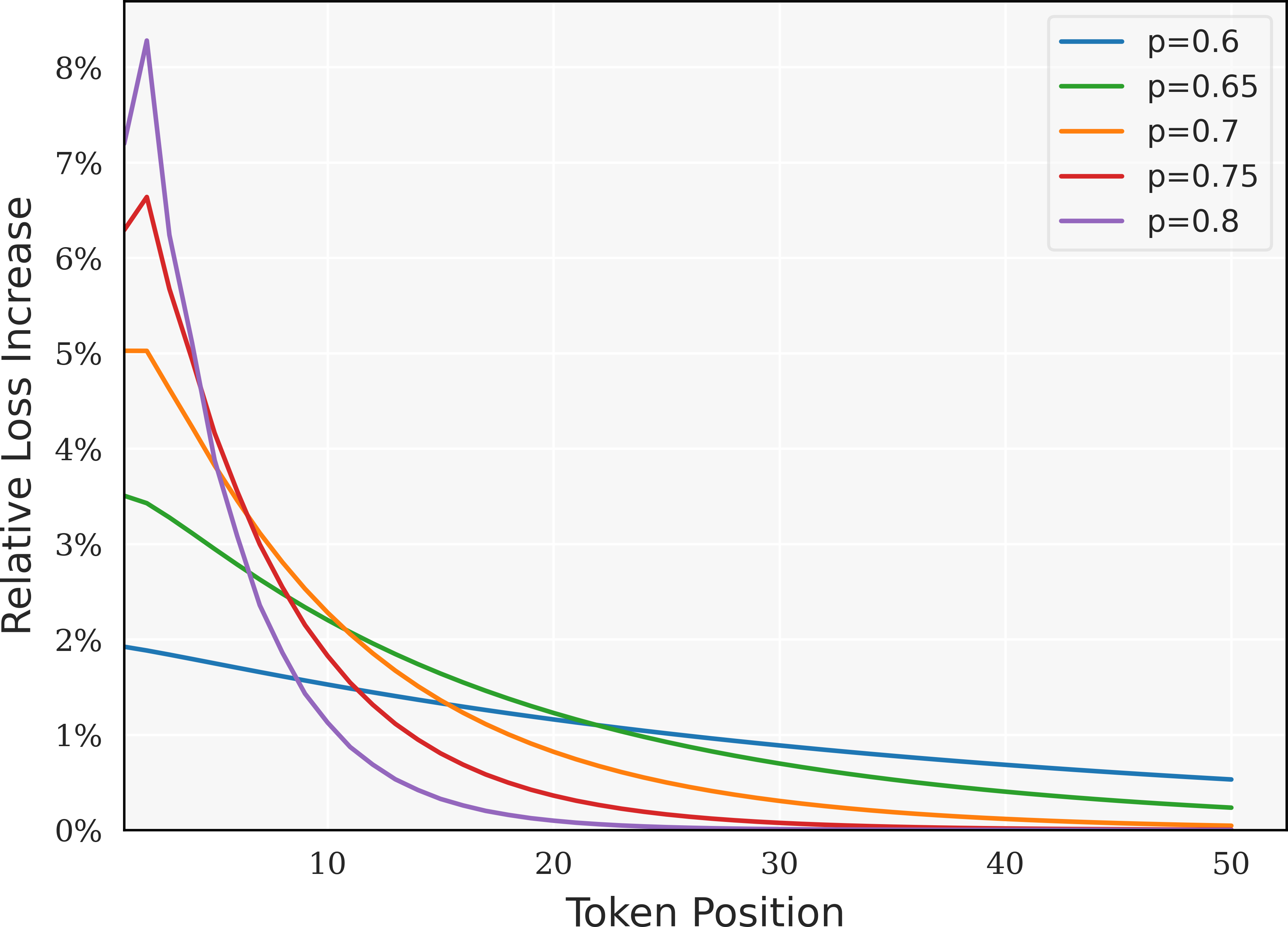}
    
    \caption{Relative increment in expected loss of model B (trained on truncated sequences) with respect to model A (trained on full sequences), as a function of token position $m \geq 1$, for different $p$.\vspace{-0.1in}}
    
    \label{fig:toy-process}
\end{figure}

For $m \geq 1$, the expected classification loss achieved by model A on token $X_m$ is given by the conditional entropy
\[
    H(X_m \mid X_{0:m}) = H(X_m \mid X_0) = -p \log p - q \log q,
\]
where $q = 1-p$.
On the other hand, if we feed a sequence of observations $(x_0, \dots, x_{m-1})$ to Model B, its prediction for the next token is equal to
\[ P \big( X_{m+1} \mid X_{1:m+1} = (x_0, \dots, x_{m-1}) \big). \]
This distribution can be computed analytically thanks to the simplicity of the process (see \Cref{seq:appendix-toy}), allowing us to determine the expected loss of model B.
The relative increase in loss from model A to model B is shown in Figure \ref{fig:toy-process} as a function of $m$. We see that model B is always worse, even under the assumption of sufficient data.
The relative loss increase always converges to $0$ as $m$ goes to infinity, but the convergence rate hinges on $p$, i.e., on how strongly the visible tokens depend on the truncated one.
This exemplifies an effect that exists also in real-world language modeling: if a key piece of information is truncated but then repeated shortly after (high $p$), then the absence of the first mention does not have a lasting impact; on the other hand, if only vaguely related information about the truncated concept is available (low $p$), then the effect of truncation lasts longer.

\section{\method}\label{sec:packing}
We propose a new method to \textit{group training data efficiently} while \textit{eliminating unnecessary truncation}, as illustrated in Figure \ref{fig:main}. Given the model's max sequence length $L$, we first segment every document into chunks that are at most $L$ tokens long. Note that this is the minimally required truncation, constrained by the context length. Then, to construct each training sequence, we select a number of document chunks to fill up as much of the $L$-token space as possible, without breaking any of them. 
The selection strategy, which we refer to as the packing algorithm, is discussed in \S\ref{subsec: packing_alg}. 
There are two challenges: first, as it's not always feasible to fill up the $L$-token sequence fully, padding tokens are used, leading to an increased number of training sequences compared to the concatenation approach. This increase necessitates more training steps per epoch. Therefore, packed sequences must be compact enough to minimize the use of padding tokens in order to retain training efficiency. Second, the algorithm must be scalable and fast enough so that it can operate on datasets of billions of documents.

\subsection{The packing algorithm}\label{subsec: packing_alg}

We formulate \method as a combinatorial optimization problem. We then present an efficient algorithm that scales linearly with data size, and show the solution achieves the same level of compactness as the usual concatenation, thus incurring a negligible loss in training efficiency. Empirically, we validate our method on large-scale pre-training datasets, specifically the RefinedWeb \cite{penedo2023refinedweb} for text, and the Stack \cite{kocetkov2022stack} for code.

Given a set of document chunks $C=\{c_1, \dots, c_{N}\}$, where $l(c)$ is the length of $c$ in tokens and $l(c_i)\leq L$, packing these chunks into training sequences is equivalent to determining a partition of $C$, denoted as $S=\{s_1, \dots, s_{M}\}$, subject to $\sum_{c\in s_i}l(c)\leq L$. A training sequence is constructed by concatenating all chunks in an $s_i$. Our goal is to find a partition $S$ of the smallest possible size, which in practical terms means generating the fewest number of training sequences.

The above optimization problem is known as \textit{the bin packing problem} \cite{bernhard2008combinatorial}, in which $N$ items of different sizes must be packed into a finite number of bins or containers, each of a fixed given capacity, in a way that minimizes the number of bins used. Computationally, the problem is NP-hard. There exist several approximation algorithms, among which \textit{First-Fit-Decreasing} (FFD) and \textit{Best-Fit-Decreasing} (BFD) are the most popular ones that strike a good balance between efficiency and accuracy. We briefly describe these heuristics in Algorithm \ref{alg:ffd}.
\begin{algorithm}[!tb]
   \caption{First/Best-Fit-Decreasing}
   \label{alg:ffd}
\begin{algorithmic}
   \STATE {\bfseries Input:} items $C=\{c_i\}_{i=1}^{N}$, bin size $L$
   \STATE Define $l(c)$: the weight of item $c$
   \STATE Define $r(b)$: the remaining capacity of bin $b$
   
   \STATE $(b_1, b_2, \dots, b_{N})\leftarrow$ Initialize empty bins 
   \STATE $SC\leftarrow$ Sort $C$ by weight in descending order
   \FOR{$c_i$ {\bfseries in} $SC$}
   \STATE Let $J=\{j|r(b_j)\geq l(c_i)\}$ %
   \begin{itemize}[noitemsep, topsep=0pt, partopsep=0pt]
        \item FFD: Find $j^*=\texttt{min}(J)$
        \item BFD: Find $j^*=\texttt{argmin}_{j\in J}r(b_j)$
   \end{itemize}
   \STATE Add $c_i$ to $b_{j^*}$
   \ENDFOR
\end{algorithmic}
\end{algorithm}

\textbf{Time Complexity } In general, both FFD and BFD take $O(N\log N)$ sorting time and $O(N\log N)$ packing time. The search step for $j^*$ is typically implemented with an $O(N)$-sized balanced binary tree that tracks all existing bins. However, in our case, notice that the sequence length is always an integer in $[1, L]$, where $L \ll N$. This reduces the sorting cost to $O(N)$ via count sort, and more importantly, allows further optimization on the packing part. Since in BFD we do not distinguish among bins with the same remaining capacity, it suffices to track the remaining capacity values instead of the actual bins, which effectively reduces the tree size to $O(L)$, and consequently the packing time to $O(N\log L)$. However, the same does not apply to FFD, because the order of bins matters.

In practice, we implemented the above fast search in BFD using a segment tree defined as follows:
\begin{itemize}[topsep=2pt, itemsep=2pt]
    \item The tree has $L$ leaf nodes.
    The value of the $i$-th leaf is $i$ if there exists at least one bin whose remaining capacity is $i$, and zero otherwise.
    Initially, all leaf nodes are set to zero, except the last one which is set to $L$.
    \item The value of every internal node is the maximum value of its children.
\end{itemize}
To find the best-fit bin, we query the tree from the root. At every internal node, we go left if the left child is no less than the item weight, and go right otherwise. We end up at a leaf node whose value is the best-fit capacity. A capacity-to-bin map is used to retrieve the best-fit bin. Finally, we update the tree to restore the two properties listed above. Please refer to Appendix \ref{appendix:bfd_figs} for a more detailed illustration.

Table \ref{table:time_complexity} presents a runtime comparison of the Optimized Best-Fit Decreasing (OBFD) algorithm against the standard First-Fit Decreasing (FFD) at 2048 context length on different data scales by up/down-sampling the RefinedWeb dataset which consists of roughly 1 billion documents.
As demonstrated, our optimized BFD saves 60\% of the running time at 1B scale, and the relative speedup (FFD/OBFD) increases logarithmically to the data size. With this efficient implementation, our method is able to scale up to even larger datasets as the asymptotic time complexity only depends \textit{linearly} on the data size.

\begin{table}[!ht]
\centering
\caption{We benchmark the running time in seconds of FFD and our optimized BFD (OBFD) on 1 million to 2 billion documents at 2048 context length. Both algorithms are implemented in Python and run on a single  CPU thread. The baseline FFD uses a similar segment tree as the optimized BFD, except having $N$ leaves that correspond to all bins. Our optimized BFD significantly improves the packing efficiency with linear scalability.} %
\resizebox{1.0\columnwidth}{!}{
\begin{tabular}{lrrrrr}
\toprule
& \multicolumn{5}{c}{Document Count} \\
 \cmidrule(lr){2-6} 
 & 1M & 10M & 100M & 1B & 2B \\
\midrule
FFD (sec) & 17 & 205 & 2,311 & 26,354 & 55,074 \\
OBFD (sec) & 10 & 106 & 1,066 & 10,816 & 22,244 \\
FFD / OBFD & 1.7 & 1.93 & 2.17 & 2.44 & 2.48 \\
\bottomrule
\end{tabular}
}
\label{table:time_complexity}
\end{table}

\begin{figure}[!ht]
    \centering
    \begin{subfigure}
    \centering
    \includegraphics[width=0.97\columnwidth]{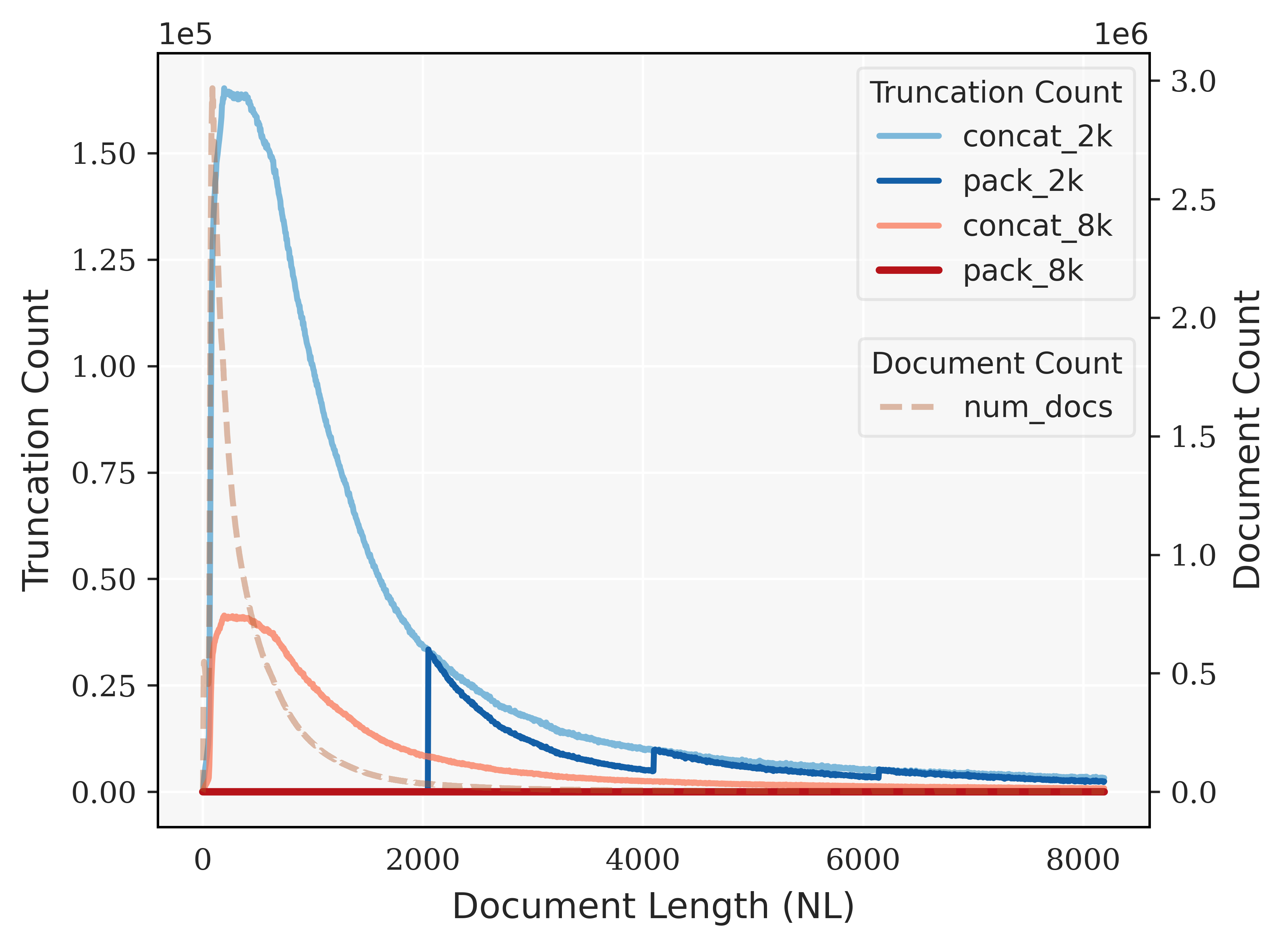}
    \end{subfigure}

    \begin{subfigure}
    \centering
    \includegraphics[width=0.97\columnwidth]{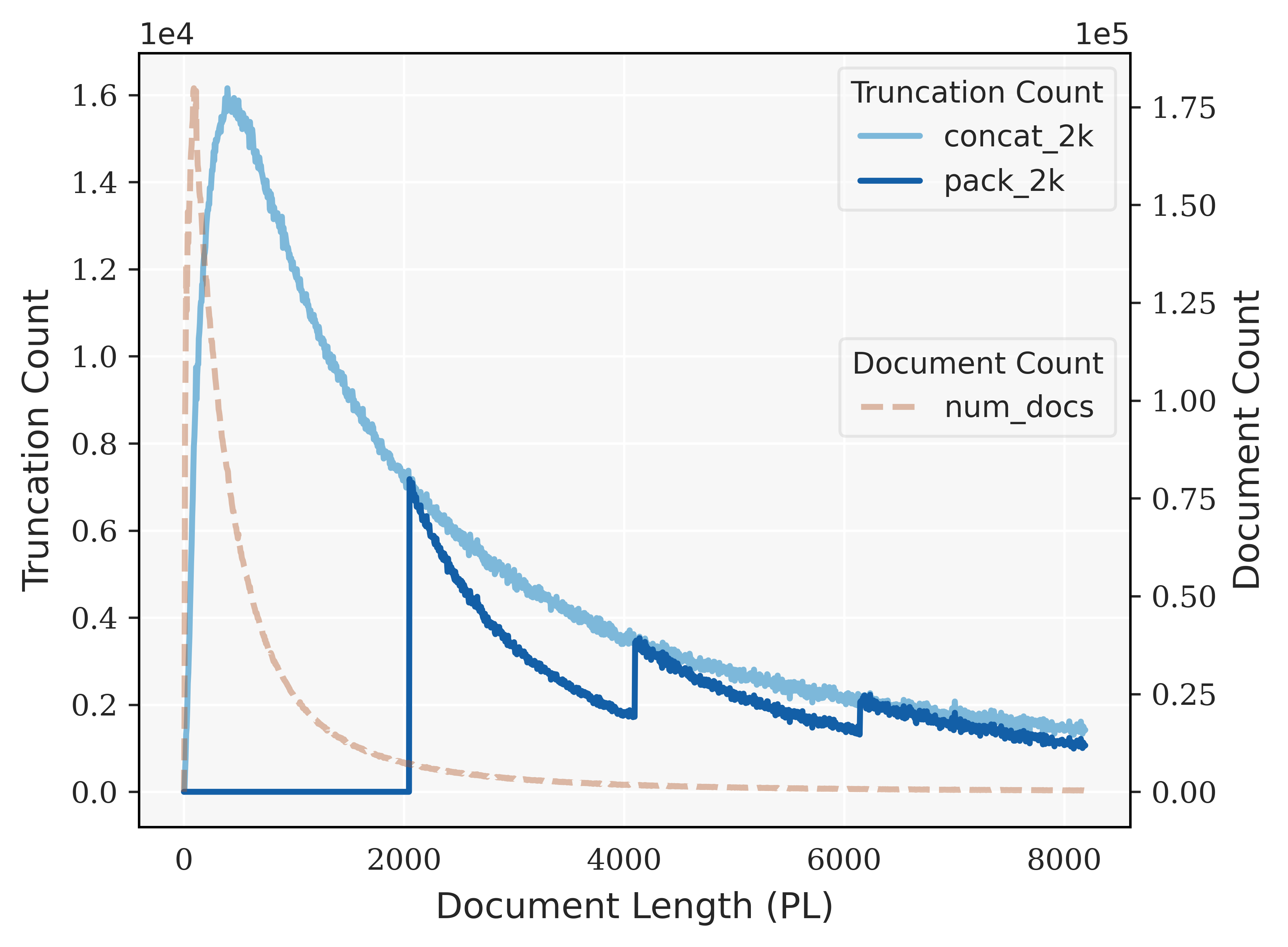}
    \end{subfigure}
    
    \vspace{-0.05in}
    \caption{Document count and truncation count for each document length, under 2k or 8k max sequence length. The number of truncations reduces significantly with \method. Top: Natural Language. Bottom: Programming Language.
    }
    \label{fig:truncation count}
\end{figure}

\textbf{Compactness } The other important perspective of packing is how compact the resulting training sequences are. Theoretically, both FFD and BFD are guaranteed to use no more than $11/9$ of the optimal number of bins asymptotically \cite{DBLP:journals/siamcomp/JohnsonDUGG74}.
However, practically, the majority of documents are short compared to context length, as shown by the dashed curves in Figure \ref{fig:truncation count}. The abundance of small-sized items makes packing much easier.
Consequently, we observe that the training sequences from packing are nearly as compact as those obtained from concatenation. As shown in Table \ref{table:compact}, \method yields no more than 0.01\% additional training sequences with either 2k or 8k context length and across NL and PL datasets. From another perspective, Table \ref{table:compact} also indicates that with \method, the amount of padding is negligibly small. The results prove that \method achieves roughly the same training efficiency as concatenation, as measured by the number of non-padding tokens processed using the same amount of compute.

\begin{table}[t]
\centering
\caption{Compactness of \method. We present the number of training sequences generated by concatenation, and the increment in number (or percentage) of sequences generated by \method with respect to concatenation. The difference between the two approaches is negligible. Therefore, \method retains the same training efficiency as concatenation. }
\resizebox{1.0\columnwidth}{!}{
\begin{tabular}{lccc}
\toprule
               & \multicolumn{2}{c}{\thead{RefinedWeb (NL)}} & \thead{Stack (PL)} \\ \midrule
Max length & 2048               & 8192              & 2048                      \\ \midrule
$Concat$  &        $2.6\times 10^8$        & $6.5\times 10^7$        & $6.4\times 10^7$                \\
$\Delta(Pack, Concat)$      & $+6253$               & $+411$                  & $+1786$                \\
$\Delta\%(Pack, Concat)$  & $0.0024\%$            & $0.00063\%$             & $0.0028\%$                  \\ \bottomrule                    
\end{tabular}
}
\label{table:compact}
\end{table}

\textbf{Truncation reduction }
We study to what extent \method alleviates the truncation problem. We count how many times each document is truncated in both NL and PL datasets, and aggregate by document length, as plotted in Figure \ref{fig:truncation count}. Note that most documents contain fewer than $L=2048$ tokens; therefore, truncation caused by concatenation predominantly occurs within this range. By completely eliminating truncation for document lengths below $L$, \method effectively preserves the integrity of an overwhelming majority of documents, limiting truncation only to longer documents where it is absolutely necessary.

\section{Experiments and Results}\label{sec: results}

To empirically validate the effectiveness of \method over concatenation, we pre-train a set of transformer language models using the same architecture as LLaMA \cite{touvron2023llama}, covering different domains, sizes, and context lengths as in Table \ref{tab:model_spec}.
We use two popular pre-training datasets in our study: the Falcon RefinedWeb dataset \cite{penedo2023refinedweb} for text,  and the Stack \cite{kocetkov2022stack} for code. Please refer to Appendix \ref{appendix:exp} for additional details on the training setup.

\begin{table}[!ht]
\centering
\vspace{-0.05in}
\caption{
Model specs. We list the number of model parameters, context length, and the number of tokens trained.
\label{tab:model_spec}
}
\resizebox{1\columnwidth}{!}{
\begin{tabular}{ccccc}
\toprule
Domain& \#Params & Context Len. & \#Tokens  \\
\midrule
Natural Language  & 13B  & 2048  & 500B \\
Natural Language  & 13B  & 8192 & 500B \\
Programming Language  & 7B  &  2048 & 300B \\
\bottomrule
\end{tabular}
}
\end{table}

\begin{table*}[!ht]
\centering
\caption{Evaluation on \textbf{Reading Comprehension} tasks. \method outperforms the concatenation baseline on almost all datasets.}
\resizebox{1.0\textwidth}{!}{
\begin{tabular}{clccccccccc}                                                  
\toprule
\multirow{2}[1]{*}{Seq Len}  & \multirow{2}[1]{*}{Method}  & Narrative QA & QuAC    & Natural Q & SQuAD &DROP & BoolQ & RACE-m  & \multicolumn{1}{c|}{RACE-h}   & \multirow{2}[1]{*}{\textbf{Average}}\\
\cmidrule(lr){3-3} \cmidrule(lr){4-4}\cmidrule(lr){5-5} \cmidrule(lr){6-6} \cmidrule(lr){7-7} \cmidrule(lr){8-8} \cmidrule(lr){9-9} \cmidrule(lr){10-10}
             &       & F1           & F1      & EM            & EM         & EM         & Acc        & Acc     & \multicolumn{1}{c|}{Acc}     &                   \\
\midrule
\multirow{2}{*}{2k}                    & Concat       & \rclow{60.38\%}          & \rclow{33.86\%}          & \rclow{39.13\%}          & \rclow{53.86\%}          & \rclow{21.47\%}          & \rclow{70.64\%}          & \rclow{50.00\%}          & \multicolumn{1}{c|}{\rclow{44.31\%}}          & 46.71\%          \\
                     & Pack         & \hspace{-1pt}\rcs{63.91\%} & \rcn{35.09\%} & \rcn{39.87\%} & \rcs{61.61\%} & \rcs{24.44\%} & \rcn{70.73\%} & \rcn{50.55\%} & \multicolumn{1}{c|}{\rcn{45.17\%}} & \textbf{48.92\%} \\
\midrule
\multirow{2}{*}{8k}                   & Concat       & \rclow{61.03\%}          & \rclow{33.68\%}          & \rclow{38.40\%}          & \rclow{59.19\%}          & \rclow{23.52\%}          & \rclow{69.33\%}          & \rcn{49.72\%} & \multicolumn{1}{c|}{\rclow{42.20\%}}          & 47.13\%          \\
                     & Pack         & \hspace{-1pt}\rcs{64.78\%} & \rcs{35.79\%} & \rcn{39.93\%} & \rcs{61.57\%} & \rcs{25.35\%} & \rcs{71.31\%} & \rclow{48.34\%} & \multicolumn{1}{c|}{\rcs{45.17\%}} & \textbf{49.03\%}
 \\      
\bottomrule
\end{tabular}
}
\label{tab:reading_comprehension}
\end{table*}

We evaluate models on a variety of downstream tasks with zero-shot and few-shot prompting.
Our findings reveal that \method improves performance in an array of tasks, most significantly in reading comprehension (+4.7\%), natural language inference (+9.3\%), context following (+16.8\%) and program synthesis (+15.0\%).\footnote{As the scale of metric varies task by task, we use relative improvement in narratives by default unless otherwise noted.} We also show that \method effectively reduces closed-domain hallucination.

Throughout this section, we denote results that are statistically significant ($p<0.05$) under a paired t-test by a superscript $\textbf{s}$, and those that are not significant by $\textbf{n}$. 
On every single task evaluated, our method either outperforms or matches the baseline. Importantly, in no case do we observe a statistically significant degradation with \method, showing that the improvement is \textit{monotonic}.

\subsection{Reading Comprehension}
Reading comprehension tests models' ability to answer questions based on information from a given passage.
We evaluate the 13B natural language models with 5-shot in-context learning on Narrative QA \cite{narrativeqa}, Natural Questions \cite{naturalqa}, SQuAD \cite{squadv2}, and DROP \cite{drop}; with 1-shot on QuAC \cite{quac}; and with zero-shot on BoolQ \cite{boolq} and RACE \cite{race}. We report F1 score or exact match (EM) for generation tasks, and accuracy for multiple-choice tasks. We adopt the few-shot examples released by HELM \cite{helm} for Narrative QA, QuAC, and Natural Questions, and randomly sample from the training split for the rest datasets. %

Results in Table \ref{tab:reading_comprehension} demonstrate the superior performance of \method in reading comprehension at both 2k and 8k context length: packing significantly outperforms concatenation in half of the settings, and shows no degradation on the rest.
Across different benchmarks, we observe: 1) between open-ended generation and multiple choice, \method generally achieves more significant gains on the former, presumably because hallucination is less of a problem in multiple choice questions where the answer space is highly constrained. 2) Among generation tasks, the improvement on  Natural Questions is minimal. We speculate that this is because these questions are designed to be answerable under both open-book and closed-book settings, which allows models to occasionally bypass the context and still provide correct answers based on their inherent parametric knowledge. 3) On all other generation tasks where the answer must be inferred from context, the improved performance aligns with our hypothesis that \method enhances models' context-awareness.

\subsection{Natural Language Inference}
\begin{table}[!th]
\centering
\vspace{-7.5pt}
\caption{\method excels over baseline in \textbf{Natural Language Inference} and \textbf{Context Following}.}
\resizebox{1.0\columnwidth}{!}{
\begin{tabular}{clcccc}
\toprule
\multicolumn{1}{c}{\multirow{3}[1]{*}{Len}} & \multicolumn{1}{c}{\multirow{3}[1]{*}{Method}}  & \multicolumn{2}{c|}{\textbf{NLI}}    & \multicolumn{2}{c}{\textbf{Context Following}}       \\\cmidrule(lr){3-4} \cmidrule(lr){5-6} 
& & \multicolumn{1}{c}{MNLI} & \multicolumn{1}{c|}{RTE}              & NQ-Swap & MemoTrap    \\
\cmidrule(lr){3-3} \cmidrule(lr){4-4} \cmidrule(lr){5-5} \cmidrule(lr){6-6}
\multicolumn{1}{c}{}                         & \multicolumn{1}{c}{}                        & ACC               & \multicolumn{1}{c|}{ACC}               & \multicolumn{1}{c}{EM}               & \multicolumn{1}{c}{ACC}                \\ \midrule
\multirow{2}{*}{2k}                          & Concat                                      & 42.78\%           & \multicolumn{1}{c|}{55.74\%}           & \multicolumn{1}{c}{45.62\%}          & \multicolumn{1}{c}{35.58\%}            \\
                                             & Pack                                        & \nls{44.33\%}     & \multicolumn{1}{c|}{\nls{60.28\%}}     & \multicolumn{1}{c}{\nls{51.03\%}}    & \multicolumn{1}{c}{\nls{41.56\%}}   \\ \midrule
\multirow{2}{*}{8k}                          & Concat                                      & 38.85\%           & \multicolumn{1}{c|}{54.66\%}           & \multicolumn{1}{c}{47.07\%}          & \multicolumn{1}{c}{37.61\%}	        \\
                                             & Pack                                        & \nls{40.60\%}     & \multicolumn{1}{c|}{\nls{59.72\%}}     & \multicolumn{1}{c}{\nls{50.04\%}}    & \multicolumn{1}{c}{\nls{40.28\%}}   \\ \bottomrule
\end{tabular}
}
\vspace{-7.5pt}
\label{tab:nqswap}
\end{table}

We evaluate models' capability in understanding dependencies between sentences through natural language inference (NLI) tasks. We use 5-shot in-context learning for Multi-NLI \cite{multinli} and RTE \cite{superglue}. As shown in Table \ref{tab:nqswap}, \method improves NLI performance by up to +9.3\%. With truncation reduction, dependency relations between sentences in training documents are better preserved, which explains the observed improvement.

\begin{table*}[!ht]
\centering
\caption{Evaluation on \textbf{Summarization} tasks. We report ROUGE for accuracy, SummaC, QAFactEval, FAVA for faithfulness, and number of generated sentences. \method generally achieves better performance except on XSUM at 2k length. $\uparrow$ indicates higher is better.}
\resizebox{1.0\textwidth}{!}{
\begin{tabular}{clccccccc}
\toprule
\multirow{2}[1]{*}{Seq Len}  & \multirow{2}[1]{*}{Method}           & \multicolumn{4}{c}{CNN/DailyMail}                                                         & \multicolumn{3}{c}{XSUM}                                                                   \\
\cmidrule(lr){3-6} \cmidrule(lr){7-9}
 & & ROUGE-1/2/L  & SummaC$(\uparrow)$ & QAFactEval$(\uparrow)$ & \#Sent.  & ROUGE-1/2/L  & FAVA$(\uparrow)$  & \#Sent. \\\midrule
\multirow{2}{*}{2k}                         & Concat               & 29.46 / 11.04 / 19.47          & 32.77\%             & 3.348        & 4.89               & \hspace{4pt}\textbf{35.42} / \textbf{13.27} / \nsigs{28.23} & \hspace{2pt}\smn{86.20\%}     & 1.02       \\
                                            & Pack                 & \hspace{4pt}\textbf{33.79} / \textbf{13.14} / \sigs{22.88}   & \sms{39.55\%}      & \hspace{3.5pt}\sigs{3.772}   & \textbf{3.13}     & 35.40  / 13.14  / 27.79          &  85.57\%    & 1.02    \\
\midrule
\multirow{2}{*}{8k}                         & Concat               & 33.92 / 13.41 / 23.18          & 48.89\%             & 4.137          & 3.72               & 33.99  / 12.65  / 26.99          & 84.50\%        & 1.02     \\
                                            & Pack                 & \hspace{4pt}\textbf{35.71} / \textbf{14.43} / \sigs{23.84} & \smn{49.65\%}    & \hspace{3.5pt}\sigs{4.224}  & \textbf{3.23}        & \hspace{4pt}\textbf{35.29} / \textbf{13.35} / \sigs{27.69} & \hspace{2pt}\smn{85.80\%}   & 1.03        \\
\bottomrule
\end{tabular}
}
\label{tab:summarization}
\end{table*}

\begin{table*}[!ht]
\centering
\vspace{-7.5pt}
\caption{Evaluation on \textbf{Commonsense and Closed-book QA} tasks. \method is slightly better on average, with notable gains on ARC-C and TriviaQA.
}
\resizebox{0.85\textwidth}{!}{
\begin{tabular}{clccccccc}
\toprule
 \multirow{2}[1]{*}{Seq Len}       &   \multirow{2}[1]{*}{Method}           & HellaSwag        & PIQA             & SIQA            & ARC-E            & ARC-C            & \multicolumn{1}{c|}{TriviaQA}         & \multirow{2}{*}{\textbf{Average}} \\\cmidrule(lr){3-3} \cmidrule(lr){4-4} \cmidrule(lr){5-5} \cmidrule(lr){6-6} \cmidrule(lr){7-7} \cmidrule(lr){8-8} 
 &    & Acc              & Acc              & Acc              & Acc              & Acc              & \multicolumn{1}{c|}{EM}               &                          \\\midrule
\multirow{2}{*}{2k}     & Concat       & \csn{75.43\%} & \csn{80.25\%} & 53.53\%          & \csn{76.14\%} & 43.94\%          & \multicolumn{1}{c|}{57.01\%}          & 64.38\%                  \\
        & Pack         & 75.31\%          & 79.98\%          & \csn{53.99\%} & 75.88\%       & \css{46.67\%} & \multicolumn{1}{c|}{\csn{57.20\%}} & \textbf{64.84\%}         \\\midrule
\multirow{2}{*}{8k}      & Concat       & 74.30\%          & 79.82\%          & 53.74\%         & \csn{74.75\%} & 43.00\%          & \multicolumn{1}{c|}{55.21\%}          & 63.47\%                  \\
        & Pack         & \csn{75.16\%}  & \css{80.63\%}    & \csn{53.89\%} & 74.33\%          & \csn{44.20\%} & \multicolumn{1}{c|}{\css{56.65\%}} & \textbf{64.14\%}\\\bottomrule        
\end{tabular}
}
\label{tab:commonsense}
\end{table*}

\subsection{Context Following}
\label{subsec:context-following}

To validate our hypothesis that excessive truncations impair factual consistency and faithfulness of generation with respect to the context, we consider special cases where the context contradicts the model's parametric knowledge and the model must follow instructions or facts in the context to answer correctly. Specifically, we evaluate with 5-shot on NQ-Swap \cite{nqswap}, a perturbed version of Natural Questions by replacing both the answer and answer mentions in the context with a different entity, and with zero-shot on MemoTrap \cite{memotrap}, where the instruction conflicts with models' memorization.
Table \ref{tab:nqswap} shows our method excels concatenation in both settings by up to +16.8\%, thereby further strengthening our hypothesis. This also suggests that \method can potentially enhance in-context learning \cite{wei2023larger},  presenting a promising avenue for future exploration.

\subsection{Summarization}

We conduct 5-shot evaluation on CNN/DailyMail \cite{cnndm1, cnndm2} and XSUM \cite{xsum}, using few-shot examples released by HELM. We follow HELM to report ROUGE scores \cite{rouge} for accuracy on both datasets, along with SummaC (zero-shot) \cite{summac} and QAFactEval \cite{qafacteval} scores for faithfulness on CNN/DailyMail. However, these two metrics are suboptimal for XSUM (see Appendix \ref{appendix:faithfulness_metrics}). Thus, we report the binary factuality scores from FAVA \cite{fava} for faithfulness evaluation on XSUM.\footnote{A summary gets a score of 1 if FAVA does not detect any hallucination (with the article as reference evidence), and 0 otherwise.}

In Table \ref{tab:summarization}, we observe improvement in all cases except on XSUM with 2k context length, where both methods perform close to each other. Models trained with \method generally obtains not only higher ROUGE scores, but also better faithfulness.
The result further strengthens our hypothesis that excessive truncation in training data is one of the reasons for hallucination. By keeping documents in their entirety to the maximum possible extent, \method effectively improves the faithfulness of generations.

Interestingly, we also find that the choice between packing and concatenation makes a difference in the number of generated sentences on CNN/DM. The prompts we used explicitly ask models to summarize the article in 3 sentences. Table \ref{tab:summarization} shows that models trained with packing do a better job at following this instruction, while models trained with concatenation tend to compose longer summaries.
\method seems to ameliorate an issue where the conventional approach causes models to ramble on, despite being prompted with a verbalized length constraint.

\subsection{Commonsense and Closed-book QA}
\label{subsec:commonsense}

We evaluate models' commonsense and world knowledge on benchmarks where answers cannot be inferred solely from context and models must rely on their parametric knowledge. We use 5-shot in-context learning for SIQA \cite{SIQA}, ARC \cite{arc}, TriviaQA \cite{triviaqa}, and zero-shot for HellaSwag \cite{hellaswag} and PIQA \cite{PIQA}. We report exact match (EM) for TriviaQA, and multiple-choice accuracy for the rest tasks.

Results are presented in Table \ref{tab:commonsense}. \method is slightly better than concatenation on average, and individually the performance can be very close on some of the datasets. Although in \S\ref{sec:prelim} we discussed that truncation may also impede knowledge acquisition during training, the impact is not uniform. Most of the standard evaluation sets focus on \textit{common knowledge}, and for a piece of knowledge to be considered as \textit{common}, it must be associated with abundant occurrences in human language. Therefore, even if the knowledge is lost due to one document getting truncated, the model still has a good chance to learn it from other complete documents carrying the same information. In contrast, \textit{tail knowledge} that appears less frequently is more vulnerable to truncation.

This conjecture is exemplified by our observation that truncation reduction results in a larger gain on ARC-C, which covers likely more tail knowledge, over ARC-E, which covers likely more common knowledge. To verify this difference, we follow \citet{kandpal2023large} to count the co-occurrences of each question-answer pair from ARC-E and ARC-C respectively, using the pre-computed Wikipedia entity map. We report the distribution of QA co-occurrence counts in Table \ref{tab:arc_evaluation}, and show that indeed the challenge set contains a larger portion of rarely co-occurring pairs. This supports our earlier hypothesis and suggests a possible reason why LLMs struggle to learn long-tail knowledge \cite{kandpal2023large}.

\begin{table}[!ht]
\centering
\vspace{-6pt}
\caption{Distribution of QA co-occurrence counts in ARC.}
\resizebox{1\columnwidth}{!}{
\begin{tabular}{lrrrrr}
\toprule
\multirow{2}{*}{ARC} & \multicolumn{4}{c}{Co-occurrence Count} \\
 \cmidrule(lr){2-5} 
 & [1, 10) & [10, 100) & [100, 1k) & [1k, $+\infty$) \\
\midrule
Easy & 9\% & 26\% & 50\% & 14\%  \\
Challenge & 15\% & 32\% & 42\% & 11\% \\
\bottomrule
\end{tabular}
}
\label{tab:arc_evaluation}
\end{table}

\subsection{Program Synthesis}
\begin{table*}[!ht]
\centering
\caption{Evaluation on \textbf{Program Synthesis} tasks. \method outperforms concatenation in terms of execution-based accuracy (Pass@$k$). More importantly, our method can significantly reduce hallucination by up to 58.3\% as measured by undefined name errors.
}
\resizebox{1.0\textwidth}{!}{
\begin{tabular}{clcccccccc}
\toprule
  \multirow{3}[1]{*}{Seq Len}                    &    \multirow{3}[1]{*}{Method}      & \multicolumn{6}{c}{\textbf{Accuracy ($\uparrow$)}}                                                       & \multicolumn{2}{c}{\textbf{Hallucination ($\downarrow$)}}                                                   \\\cmidrule(lr){3-8} \cmidrule(lr){9-10} 
  & & \multicolumn{3}{c}{HumanEval} & \multicolumn{3}{c}{MBPP}   & \multicolumn{1}{c}{HumanEval} & \multicolumn{1}{c}{MBPP} \\
  \cmidrule(lr){3-5} \cmidrule(lr){6-8} \cmidrule(lr){9-9} \cmidrule(lr){10-10}
                    &           & Pass@1           & Pass@10           & Pass@100          & Pass@1           & Pass@10           & Pass@100         & Undef. Name     & Undef. Name          \\\midrule
\multirow{2}{*}{2k} & Concat    & 17.54\%          & 25.91\%           & 35.28\%           & 22.60\%          & 42.49\%           & 59.46\%          & 5.10\%          & 9.52\%               \\
                    & Pack      & \pls{18.32\%}    & \pls{28.96\%}     & \pls{40.57\%}     & \pls{23.48\%}    & \pls{45.58\%}     & \pls{62.93\%}    & \pls{2.41\%}    & \pls{3.97\%}      \\\bottomrule

\end{tabular}
}
\label{tab:code}
\end{table*}

We evaluate the 7B programming language models on HumanEval \cite{codex} and MBPP \cite{mbpp} for zero-shot code generation. Following standard practice, we generate 200 samples per problem, and report Pass@$k$ with $k=1,10,100$ for functional correctness.

Besides, a crucial and great aspect of evaluation with programming language is that we can detect hallucination \textit{accurately} without relying on ML-based models which can be error-prone. We resort to program analysis as a more reliable alternative for hallucination detection. Following \cite{ding-etal-2023-static}, we identify \textit{undefined name} errors using static analysis,
and report the percentage of generations with at least one such error as the hallucination metric.

As shown in Table \ref{tab:code}, our method both improves Pass@$k$ (+15.0\% for Pass@100 on HumanEval and +5.8\% on MBPP), and reduces undefined name errors significantly by up to 58.3\%. With \method eliminating most truncations in training inputs (cf. Table \ref{fig:truncation count}), models are exposed to fewer partial code segments that contain undefined names. Consequently, hallucination is suppressed in model-generated code. Besides that, \method also benefits functional correctness as reflected in Pass@$k$ improvement, which we believe is a combined effect of reduced hallucination and a better understanding of programming logic by learning from complete code examples.

\section{Related Work}\label{sec:related_work}

\textbf{Pre-training Data} 
Pre-training data is pivotal to the quality of language models. There has been multiple high-quality pre-training datasets that were made publicly available, e.g., C4 \cite{t5}, Pile \cite{gao2020pile}, RefinedWeb \cite{penedo2023refinedweb}, RedPajama \cite{together2023redpajama}, and the Stack \cite{kocetkov2022stack,lozhkov2024starcoder}. On top of these, multiple papers (e.g., \cite{lee2022deduplicating,marion2023less,chen2023data,chowdhery2023palm,touvron2023llama,raffel2020exploring}) propose various filtering strategies to improve data quality. Our work broadly applies on top of these pre-training datasets.

\textbf{Data grouping in language model training}
Recent transformer language models have adopted different strategies to group training data into batched sequences in order to tackle the variable document length problem. For encoder-only models, the choice of data formatting was first studied in RoBERTa \cite{roberta}, which shows that concatenating sentences from more than one documents in the same training sequence results in very little performance degradation. \citet{bert_packing} proposed an approximation-based combinatorial packing method to accelerate BERT training \cite{bert}, yet without improving downstream performance. 
It is worth mentioning that document truncation is less of a concern for encoder models for two reasons: first, they are usually trained on relatively short text spans of 128-512 tokens that only respect sentence boundary. In such case, document-wise truncation is inevitable given the limited context size. Second, they are not intended for open-ended generation, and thus, hallucination is not an issue.

Decoder-only language models have predominantly adopted the concatenate-then-split strategy to maximize training efficiency \cite{gpt3,palm,gopher,opt,llama2,bloom}. Very recently, \citet{shi2024incontext} proposed to concatenate semantically relevant documents into the same training sequence, which yields notable improvement on downstream tasks. Nevertheless, the method as a variant of concatenation still suffers from excessive truncation, and it is orthogonal (and possibly complimentary) to our method.

\textbf{Integration with LLM Training Framework} 
Best-fit Packing operates on data level and can be handled as an offline process. Thus, it does not require any change in the training implementation and can be integrated into common distributed training frameworks like Megatron-LM \cite{shoeybi2019megatron} or DeepSpeed \cite{rasley2020deepspeed} easily. 

\textbf{Hallucination in Language Generation}
With the rapid development of generative language models of large scale, hallucination has attracted increased attention as it can hinder performance and mislead users with fabricated facts \cite{hallu_survey}. Various approaches have been proposed to tackle this problem, including retrieval augmented generation \cite{peng2023check, kang2023ever}, prompt engineering \cite{si2023prompting, ji-etal-2023-towards}, context-aware decoding \cite{shi2024incontext}, and supervised finetuning \cite{tian2023finetuning}. However, hallucination mitigation during the pre-training stage has largely been overlooked, and we are among the first to explore in this direction.

\section{Conclusion}\label{sec:conclusion}

The prevalent concatenate-then-split approach of data grouping in language model training inevitably results in fragmentation of documents.  We show that this truncation effect undermines models' ability to follow the context, and even worse, makes models more prone to hallucination. Motivated by these, we propose \method, a new data grouping method that maximally preserves the entirety of individual documents. The algorithm is scalable for datasets of billions of documents, and maintains the same level of compactness as concatenation. Empirically, we demonstrate the effectiveness of truncation reduction by comparing models trained with different data grouping strategies at various scales across both text and code.
Specifically, we show that by eliminating unnecessary truncations, \method excels in a broad range of tasks without compromising performance on others. Additionally, it effectively reduces closed-domain hallucination in language generation. While the experiments conducted in this paper have primarily focused on the pre-training stage, \method is broadly applicable to the finetuning stage as well. Our work contributes to the ongoing efforts in developing more effective and reliable language models.

\section*{Impact Statements}
This paper highlights a critical and fundamental issue of excessive truncation that broadly exists in the contemporary LLM training practice. We propose \method, a novel method that mitigates the issue with wide applicability in LLM training. We demonstrate that it improves language modeling while reducing hallucination. However, it is crucial to acknowledge that no method can completely eliminate the risk of generating false information or hallucinating. We encourage readers to refer to \citet{weidinger2021ethical} for an in-depth discussion on societal impact and potential risks associated with LLMs.
\section*{Acknowledgements}
We would like to express our sincere gratitude to Carson Klingenberg, Wasi Ahmad, and colleagues at AWS AI Labs for their help and valuable feedback.  We are also deeply grateful to the anonymous reviewers whose insightful and constructive comments have significantly enhanced the quality of this work.

\newpage
\bibliography{paper}
\bibliographystyle{icml2024}

\newpage
\appendix
\onecolumn
\section{Derivation of Model Accuracy on the Toy Process}
\label{seq:appendix-toy}

We now describe the computations needed to compute the accuracy of models A and B on the toy process described in \S\ref{sec:analytical}.

Let $x = (x_0, \dots, x_{m-1})$ be a sequence of observations, for a fixed $m \geq 1$.
The next token $X_m$ is distributed as a Bernoulli variable, taking the value $x_0$ with probability $p$ and $1-x_0$ with probability $q = 1-p$.
This is exactly the distribution predicted by model A.
The expected classification loss of model A on the $m$-th token is therefore given by the entropy of a Bernoulli distribution, and it does not in fact depend on $m$ or on the value of $x_0$ (because the entropy remains the same if we swap $p$ and $q=1-p$).

Model B, however, ``believes'' that the observations $x$ are part of a sequence $(\tilde x_0, x_0, \dots, x_{m-1})$ drawn according to the process $(X_n)_{n \in \mathbb{N}}$.
We can then think of the inference of model B as consisting of two steps: first, predict the probability distribution of the hidden token $\tilde x_0$; then, predict the probability distribution of the next token in the sequence, which is $X_{m+1}$ from the point of view of model B. In formulas:
\begin{align}
    & P \Big(X_{m+1}=0 \mid X_{1:m+1} = x \Big) \nonumber\\
    &= \sum_{\tilde x_0 \in \{0,1\}} P \Big( X_0 = \tilde x_0 \mid X_{1:m+1} = x \Big) \cdot P \Big(X_{m+1}=0 \mid X_0=\tilde x_0\Big) \nonumber\\
    &= P \Big( X_0 = 0 \mid X_{1:m+1} = x \Big) \cdot p
    \,+\, P \Big( X_0 = 1 \mid X_{1:m+1} = x \Big) \cdot q.
    \label{eq:model-b-prediction}
\end{align}
The distribution of the hidden token can be computed using Bayes' rule:
\begin{align*}
  &P \Big( X_0 = 0 \mid X_{1:m+1} = x \Big) \\
  &=
  \frac{P \Big( X_{1:m+1} = x \mid X_0 = 0 \Big) \cdot P \Big(X_0=0\Big)}{\displaystyle \sum_{\tilde x_0 \in \{0, 1\}} P \Big( X_{1:m+1} = x \mid X_0 = \tilde x_0 \Big) \cdot P\Big(X_0=\tilde x_0\Big)} \\
  &= \frac{p^k q^{m-k} \cdot \frac12}{p^k q^{m-k} \cdot \frac12 + p^{m-k} q^k \cdot \frac12}
  = \frac{p^{k} q^{m-k}}{p^{k} q^{m-k} + p^{m-k} q^{k}}
\end{align*}
where $k$ is the number of zeros in the sequence $x$.
We can substitute back in \eqref{eq:model-b-prediction} and obtain an explicit expression for the prediction of model B:
\begin{equation}
  P \Big(X_{m+1}=0 \mid X_{1:m+1} = x \Big) =
  \frac{p^{k+1} q^{m-k} + p^{m-k} q^{k+1}}{p^{k} q^{m-k} + p^{m-k} q^{k}}.
  \label{eq:model-B-prediction2}    
\end{equation}
The expected classification loss of model B (for the given sequence of observations $x$) is the cross entropy between the true distribution (a Bernoulli distribution that assigns probability $p$ to $x_0$ and probability $q$ to $1-x_0$) and the distribution predicted by model B, which is given by \eqref{eq:model-B-prediction2}:
\begin{equation}    
  \text{loss}_B(x) =
  -p^{1-x_0} q^{x_0}
  \log \frac{p^{k+1} q^{m-k} + p^{m-k} q^{k+1}}{p^{k} q^{m-k} + p^{m-k} q^{k}}
  - p^{x_0} q^{1-x_0} \log \frac{p^{k} q^{m-k+1} + p^{m-k+1} q^{k}}{p^{k} q^{m-k} + p^{m-k} q^{k}}
  \label{eq:mode-B-loss}
\end{equation}
where the first term corresponds to the outcome $0$ while the second term corresponds to the outcome $1$.
Note that the right-hand side of \eqref{eq:mode-B-loss} only depends on $x_0$ and $k$, so we can call it $\text{loss}_B(x_0, k)$.
Finally, the expected loss of model B is the average of $\text{loss}_B(x_0, k)$ over all possible observations $x$:
\[
  \text{loss}_B = \frac12 \sum_{x_0 \in \{0, 1\}}
  \sum_{h=0}^{m-1} \binom{m-1}{h} (p^{1-x_0} q^{x_0})^h (p^{x_0} q^{1-x_0})^{m-1-h} \text{loss}_B(x_0, h+1-x_0),
\]
where $h$ indicates the number of zeros in the subsequence $(x_1, \dots, x_{m-1})$.

\section{An Illustration of optimized BFD Algorithm}
\label{appendix:bfd_figs}
We illustrate the proposed optimized Best-Fit-Decreasing algorithm with figures. At a high-level, we maintain three data structures: (i) a \textit{bin-to-items table} which tracks the current assignment of items to every bin; (ii) a \textit{space-to-bins table} that allow us to retrieve a bin given a certain remaining space; (iii) a \textit{segment tree} that enables us to find the best-fit capacity in $O(\log L)$ time. In this example, we assume the max sequence length (or max bin capacity) is 8.

\textbf{Initialization: }Both the \textit{bin-to-items table} and the \textit{space-to-bins table} are empty at the beginning. All nodes in the \textit{segment tree} are zero except for the right most path along which the nodes have value 8, because we only have empty bins of max capacity at the very beginning.

\begin{figure*}[ht]
    \centering
    \includegraphics[trim=0 0 0 0, clip, width=0.8\textwidth]{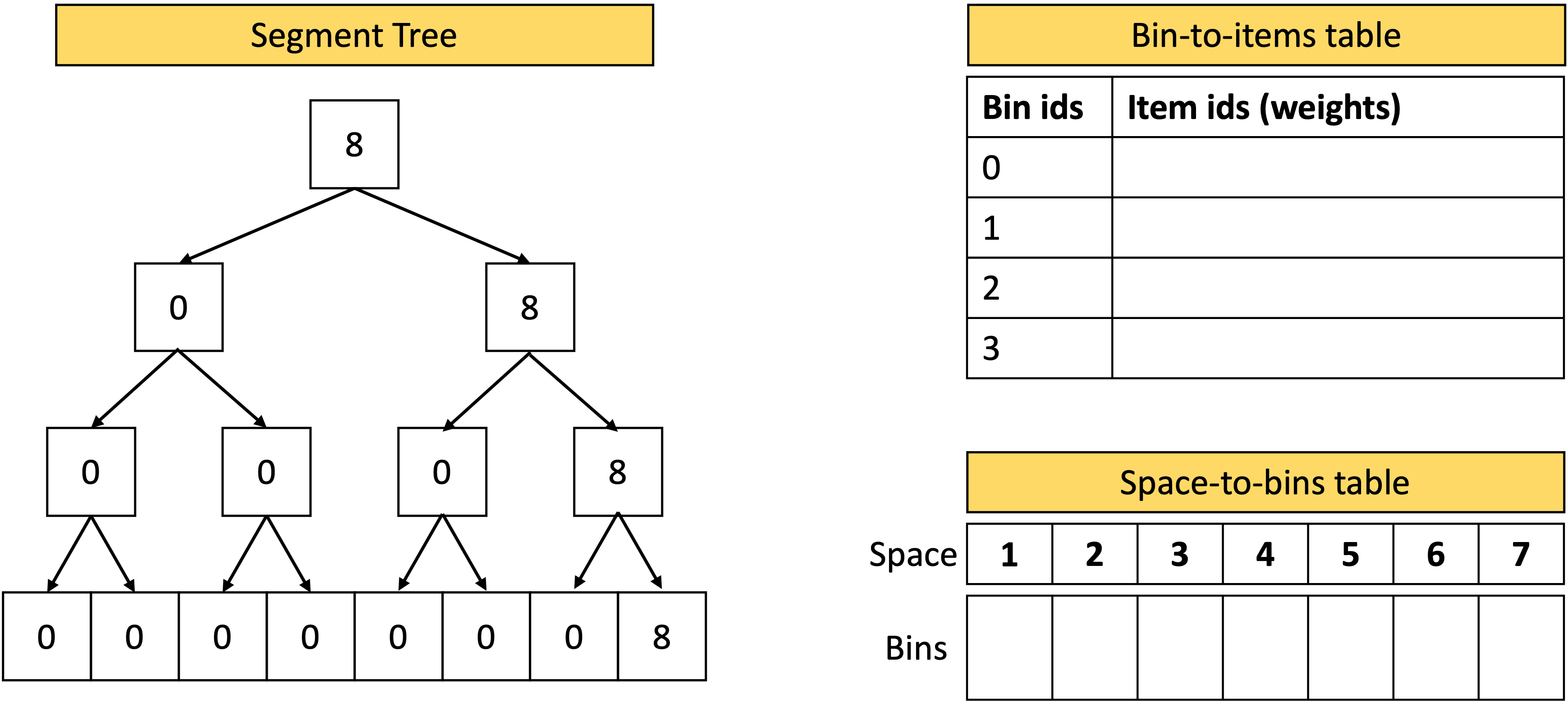} 
    \caption{Initialization}
    \label{fig:bfd0}
\end{figure*}

\textbf{A running example: }Below we demonstrate how to pack a new item given an intermediate state of the algorithm. Assume that we have packed 4 items whose weights are 8, 6, 6, 4, and arrive at the following state. Now we are about to pack an incoming item of weight 3.

\begin{figure*}[ht]
    \centering
    \includegraphics[trim=0 0 0 0, clip, width=0.8\textwidth]{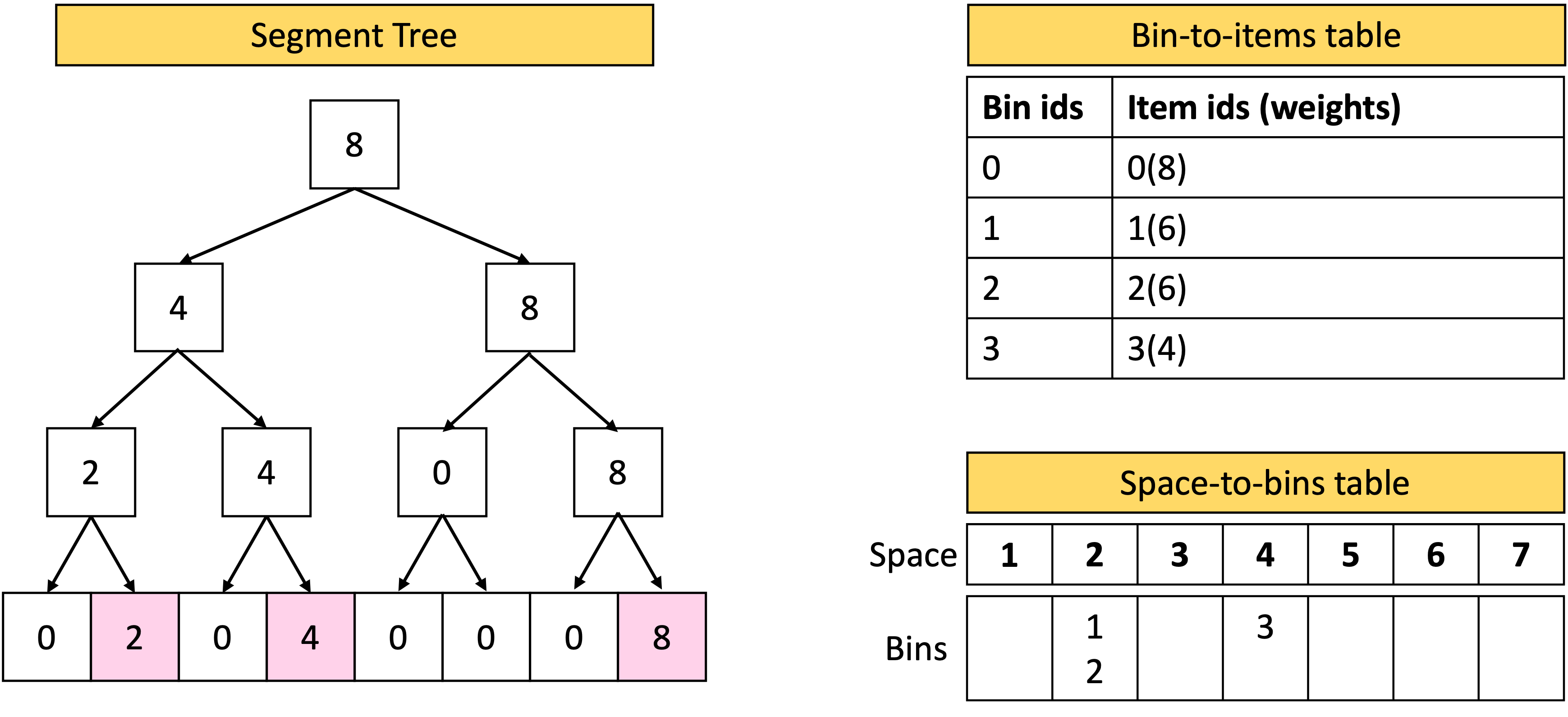} 
    \caption{State after packing four items of weight 8, 6, 6, 4. The \textit{bin-to-item table} shows the 0\textsuperscript{th} item (of weight 8) has been placed in bin 0, and the 1\textsuperscript{st} item (of weight 6) in bin 1, and etc.. The \textit{space-to-bins} table shows bin 1 and bin 2 each has 2 space left after packing an item of weight 6, and bin 3 has 4 space left. In the \textit{segment tree}, since we have bins with 2, 4, or 8 space left, the 2\textsuperscript{nd}, 4\textsuperscript{th}, and 8\textsuperscript{th} leaves from left to right are assigned value 2, 4, 8, respectively. All other leaves are zero. The internal nodes are recursively updated to be the maximum of their children.}
    \label{fig:bfd1}
\end{figure*}

\begin{figure*}[ht]
    \centering
    \includegraphics[trim=0 0 0 0, clip, width=0.8\textwidth]{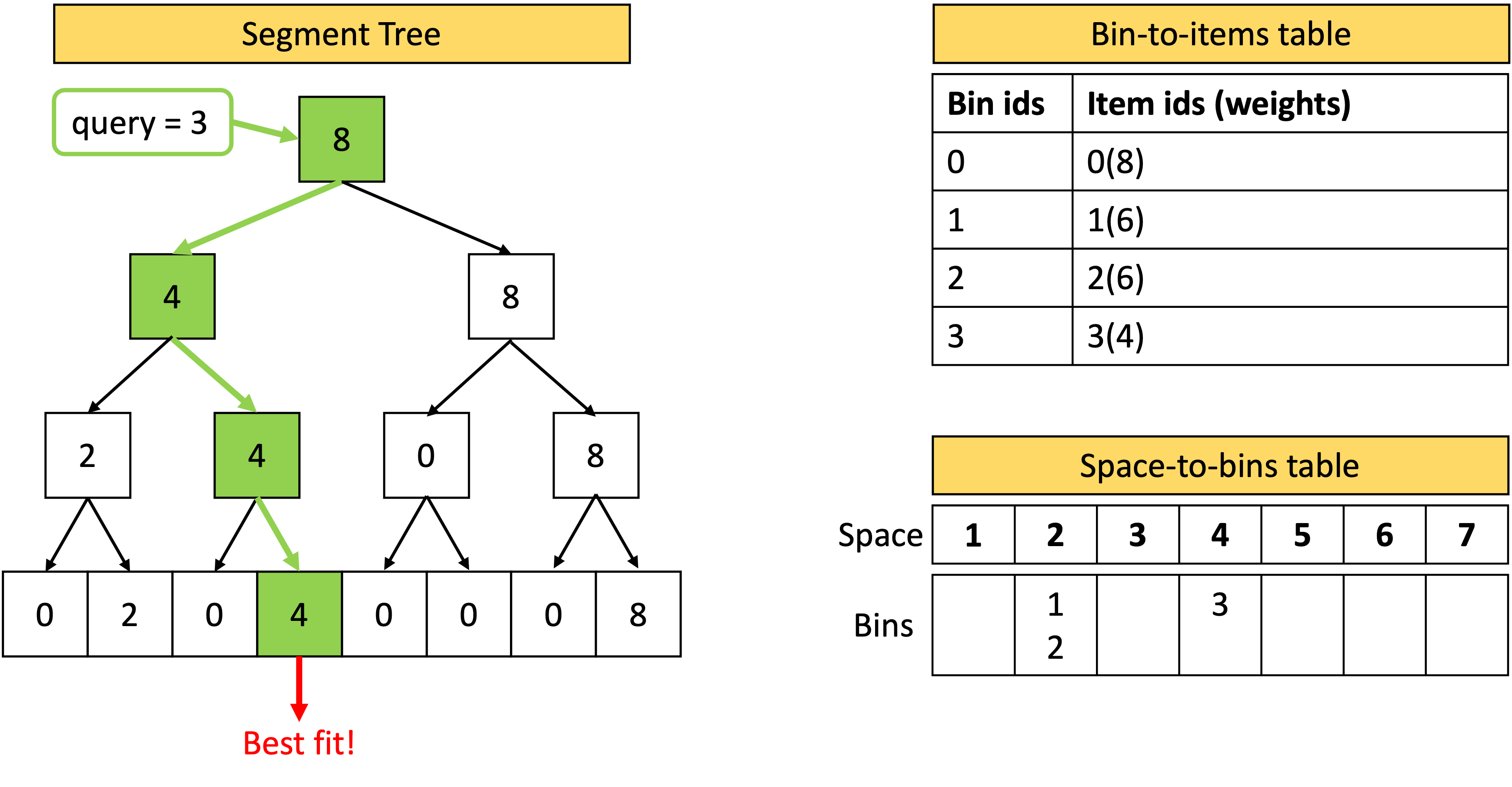} 
    \caption{To pack a new item of weight 3, the first step is to find the best-fit capacity using the segment tree. We query the tree from the root. At every internal node, we go left if the left child is no less than the item weight, and go right otherwise. We end up at a leaf node whose value is the best-fit capacity (4 in this case).}
    \label{fig:bfd2}
\end{figure*}

\begin{figure*}[!htpb]
    \centering
    \includegraphics[trim=0 0 0 0, clip, width=0.8\textwidth]{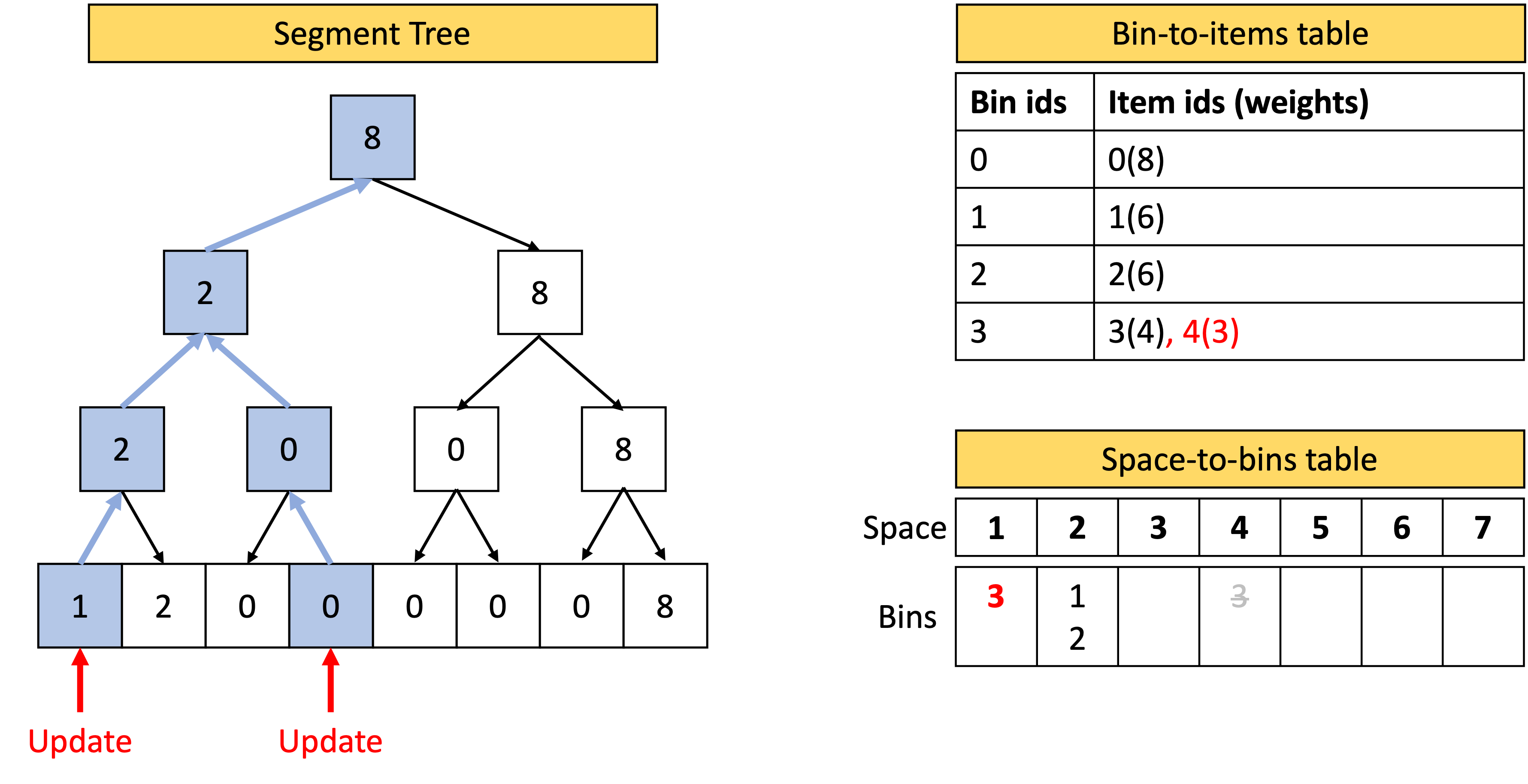} 
    \caption{We retrieve a bin with 4 space left, which is bin 3. This new 5\textsuperscript{th} item (of weight 3) is then placed in bin 3. We update the \textit{bin-to-items table} and \textit{space-to-bins} table accordingly. Finally, we update the \textit{segment tree} recursively from the bottom up to restore the two properties stated in \S\ref{sec:packing}: (i) The value of the $i$-th leaf is $i$ if there exists at least one bin whose remaining capacity is $i$, and zero otherwise; (ii) The value of every internal node is the maximum value of its children. This concludes one iteration of packing.}
    \label{fig:bfd3}
\end{figure*}

\section{Additional Details on Experiments}

\subsection{Training Details}
\label{appendix:exp}

We use the same model architecture as LLaMA 7B/13B \cite{touvron2023llama}. In case a sequence contains multiple documents, we follow the standard practice to allow attention to cross document boundaries when using concatenation \cite{gpt3, palm, llama2}, but mask that out when using \method, so that \method is equivalent to treating each document as a standalone training sample, albeit in a much more compact way. Thanks to the relative nature of rotary positional embeddings (RoPE) \cite{rope}, we do not adjust position ids from model input. We perform additional ablations on cross-document attention in \S\ref{subsec:ablation}.

We train all models with the AdamW optimizer \cite{adamw}. We use a learning rate of 3e-4 with a cosine learning rate scheduler, and warm up over the first 3,000 steps. The global batch size is 2M tokens. We use FlashAttention2 \cite{flashattention2} to accelerate training. We find the implementation of document-wise block attention in FlashAttention2 does not yield perceivable overhead compared with the standard full-sequence attention. All models were trained on a cluster of 256 A100 GPUs. 

The data pipeline for \method is implemented as follows. In the offline stage, we tokenize the raw data into document chunks up to max sequence length, and run BFD to get chunk indices for each training sequence. During training (the online stage), for every sequence, we fetch the corresponding document chunks through the chunk indices, and build the training sequence on the fly. This saves the expensive cost of concatenating individual chunks on disk.

For the Stack dataset, we use a subset of 7 popular programming languages: Python, Java, C\#, JavaScript, TypeScript, C, and C++.

\subsection{Faithfulness Metrics for Summarization Tasks}
\label{appendix:faithfulness_metrics}
We do not use SummaC \cite{summac} and QAFactEval \cite{qafacteval} for faithfulness evaluation on XSUM for two reasons. 
First, SummaC assumes that every sentence in a faithful summary can be entailed from another sentence in the article. This generally does not apply to XSUM which requires summarizing the article in one sentence. If this only sentence is an implication of just one sentence from the article, then the coverage of the summary will be very limited.
Second, from what has been reported in literature \cite{qafacteval}, both methods show lower accuracy on a summary inconsistency detection benchmark constructed from XSUM (namely XSF \cite{xsf}), compared to other benchmarks from CNN/DailyMail.

On CNN/DailyMail, we do not report the FAVA binary score because it does not take into consideration the length of the summary. The metric is biased towards shorter generations that have less chance to make a hallucination. As noted in Table \ref{tab:summarization}, there is a significant difference between the number of sentences generated by models trained with concatenation and with packing. On the contrary, on XSUM, all models are able to follow the instruction to generate 1-sentence summaries.

In general, existing methods for evaluating faithfulness of summary have certain limitations, largely due to the complexity of human language. In contrast, hallucination detection for code, which relies on program analysis, tends to be more reliable.

\subsection{Ablation Study on Cross-Document Attention}\label{subsec:ablation}

\method treats each segment as individual and thus masks out cross-document attention in training (Appendix \ref{appendix:exp}). To study the impact of this choice and to highlight the importance of our packing method, we pre-train a 2k NL model with concatenation and cross-document attention mask enabled. In Table \ref{tab:ablation}, we compare the three models by perplexity on a validation set, and performance in downstream tasks. Note that every sequence in the validation set contains only one document chunk. We find that though perplexity gain can be attributed to the attention mask, \method is critical for achieving optimal performance in downstream tasks. In particular, using attention mask on top of the conventional concatenation approach even degrades the model's performance in summarization.

\begin{table}[!ht]
\centering
\caption{Ablation results for cross-document attention. We additionally train a 2k-length model with concatenation and cross-document attention mask. We report perplexity on validation set (\textbf{PPL}), and average performance in reading comprehension (\textbf{RDC}), natural language inference (\textbf{NLI}), context following (\textbf{CTX}), summarization (\textbf{SUM}, in ROUGE-2), and commonsense (\textbf{CMS}).}
\begin{tabular}{lccccccc}
\toprule
Method & \multicolumn{1}{c}{\thead{Attn.\\Mask}} & PPL  & \multicolumn{1}{c}{\thead{RDC}} & NLI & CTX & SUM & CMS \\ \midrule
Concat & $\xmark$       & 9.64             & 46.71\%           & 49.26\%             & 40.60\%          & 12.16              & 64.38\%            \\
Concat & $\cmark$       & \textbf{9.53}    & 47.92\%           & 50.35\%             & 42.41\%          & 11.79              & 64.77\%            \\
Pack   & $\cmark$       & \textbf{9.53}    & \textbf{48.92\%}  & \textbf{52.31\%}    & \textbf{46.30\%} & \textbf{13.14}     & \textbf{64.84\%}   \\ \bottomrule
\end{tabular}
\label{tab:ablation}
\end{table}

\end{document}